\documentclass[12pt]{llncs}

\usepackage{times}

\usepackage{mathtools} 
\usepackage{changepage}
\usepackage{enumitem} 
\usepackage{xcolor} 
\usepackage{amssymb}              
\usepackage{caption} 
\newcommand{\exact}{\textsf{Exct}}
\newcommand{\pve}{\textsf{pos}}
\newcommand{\nve}{\textsf{neg}}

\newcommand{\hookrightarrowR}[1]{\ensuremath{\hookrightarrow^{#1}}}

\newcommand{\Rp}{\ensuremath{R_{\mathbf{p}}}}

\newcommand{\inAPA}{\ensuremath{\dot{\in}_\delta}}
\newcommand{\hide}[1]{} 
\newcommand{\lol}[1]{\ensuremath{\multimap^{\!\!\!\!\!\!\!#1}}}  
\newcommand{\orC}{\textsf{or}}
\newcommand{\andC}{\textsf{and}}

\begin{document}
     \title{
          Abstract Argumentation / Persuasion / Dynamics} 

\author{Ryuta Arisaka
    \and
      Ken Satoh\institute{National Institute of Informatics,
       Chiyoda, 
        Japan\\ email: ryutaarisaka@gmail.com, ksatoh@nii.ac.jp} 
    }

    \maketitle
    \bibliographystyle{plain}  
\begin{abstract} 
The act of persuasion, a key component in rhetoric argumentation, 
may be viewed as a dynamics modifier. 
We extend Dung's frameworks with acts of persuasion among agents, 
and consider interactions among attack, persuasion and defence 
that have been largely unheeded so far. 
We characterise basic notions of admissibilities in this framework, 
and show a way of enriching them through, effectively, CTL (computation tree logic) 
encoding, which also permits importation of the theoretical results 
known to the logic into our argumentation frameworks. 
Our aim is to complement the growing interest in coordination of 
static and dynamic argumentation.
\end{abstract} 
\section{Introduction}       
An interesting component of  
rhetoric argumentation 
is persuasion. We may code an act of it 
into  
$\fbox{\text{A}:\textcircled{\small $a_1$}} \dashrightarrow \fbox{\text{B}:\textcircled{\small $a_2$}} 
\lol{a_1} 
\fbox{\text{B}:\textcircled{\small $a_3$}}$ with the following 
intended meaning: 
some agent A's argument $a_1$ persuades 
an
agent B into holding $a_3$; B, being persuaded, drops $a_2$. 
There can be 
various reasons for the persuasive act. 
It may be that A 
is a great teacher wanting to correct some  
inadvisable norm of B's, or perhaps   
A is a manipulator who benefits if $a_2$ 
is not present. Persuasion is popularly observed in social forums 
including YouTube and Twitter, and methods to represent it will help understand   
 users' views on topics accurately. Another less pervasive form 
of persuasion is possible: 
$\fbox{\text{A}:\textcircled{\small $a_1$}} \multimap \fbox{\text{B}:\textcircled{\small $a_3$}}$ 
in which A persuades B with $a_1$ into expressing 
$a_3$ but without conversion. In either of the cases, persuasion  
acts as a dynamics modifier in rhetoric argumentation, allowing 
some argument to appear and disappear. \\
\indent  Of course - and this is one 
highlight of this paper  - these acts will not be 
successful if $a_1$ is detected to be not a defensible argument: 
we may have  
$\fbox{\text{C}:\textcircled{\small $a_3$}} \rightarrow 
\fbox{\text{A}:\textcircled{\small $a_1$}} \multimap \fbox{\text{B}:\textcircled{\small $a_2$}}$ 
where $a_3$ attacks $a_1$.  
Suppose now that B is aware of $a_3$, then B can defend 
against  A's persuasion due to $a_3$'s attack on $a_1$.   
B is not persuaded into holding $a_2$ in such a case. 
We will care for the interactions between attack, persuasion and 
defence. \\
\indent While AGM-like 
argumentation framework revisions defining 
a class of argumentation frameworks to result from 
an initially given argumentation framework and 
an input (which could be argument(s), attack(s)
or both), and persuasion in the context of (often two-parties) 
dialogue games, are being studied, there are very few 
studies in the literature 
that pursue coordination of statics and dynamics. 
One exception is the dynamic logic for programs adapted 
for argumentation by Doutre {\it et al.} 
\cite{Doutre14,Doutre17}, which is rich  
in expressiveness with non-deterministic operations, tests,
sequential operations. 
Bridging dynamics and statics is important for 
detailed and more precise analysis of rhetoric argumentation. 
So far, however,  
the above-said interaction between attack, persuasion and defence 
has been largely unheeded. We first of all fill the gap 
by developing an abstract persuasion argumentation, 
an extension to Dung's argumentation theory \cite{Dung95}. 
We formulate the notion of static admissibility for our 
theory, and then show a way of diversifying it into other types of 
admissibilities through, effectively, CTL (computation tree logic) embedding.   
\subsection{Example situations}     
 \begin{center} 
\includegraphics[scale=0.11]{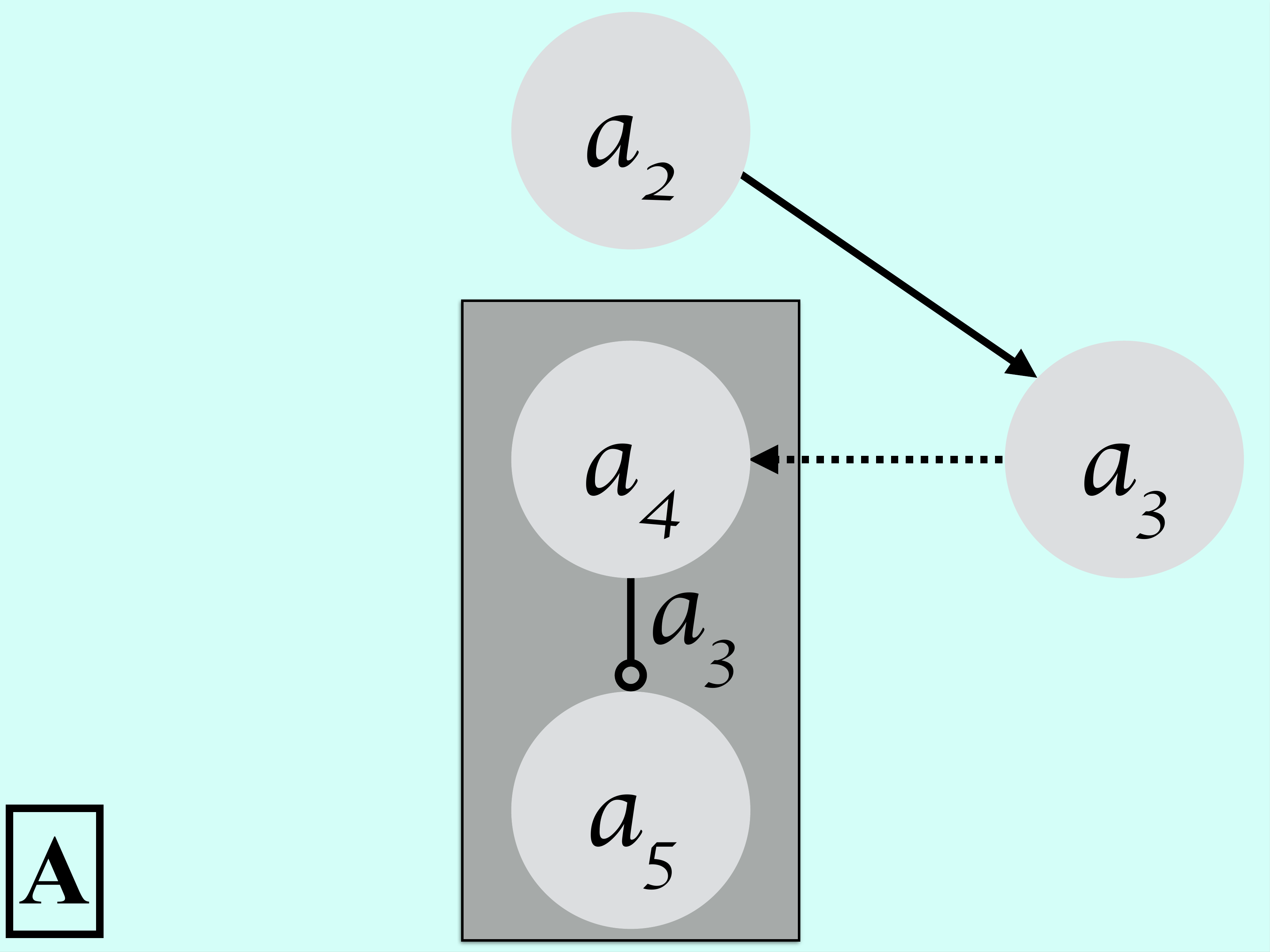} 
\includegraphics[scale=0.11]{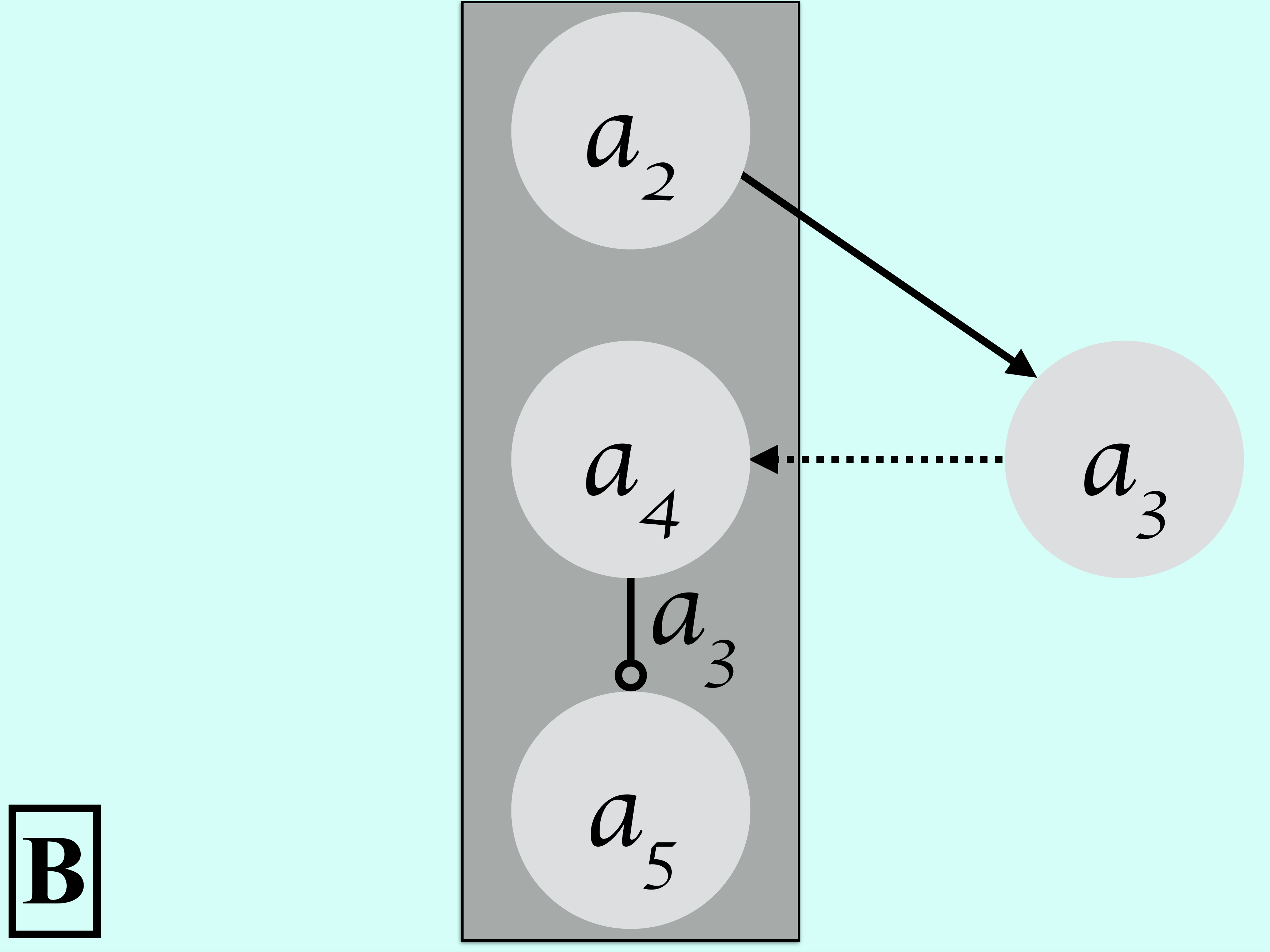} 
\includegraphics[scale=0.11]{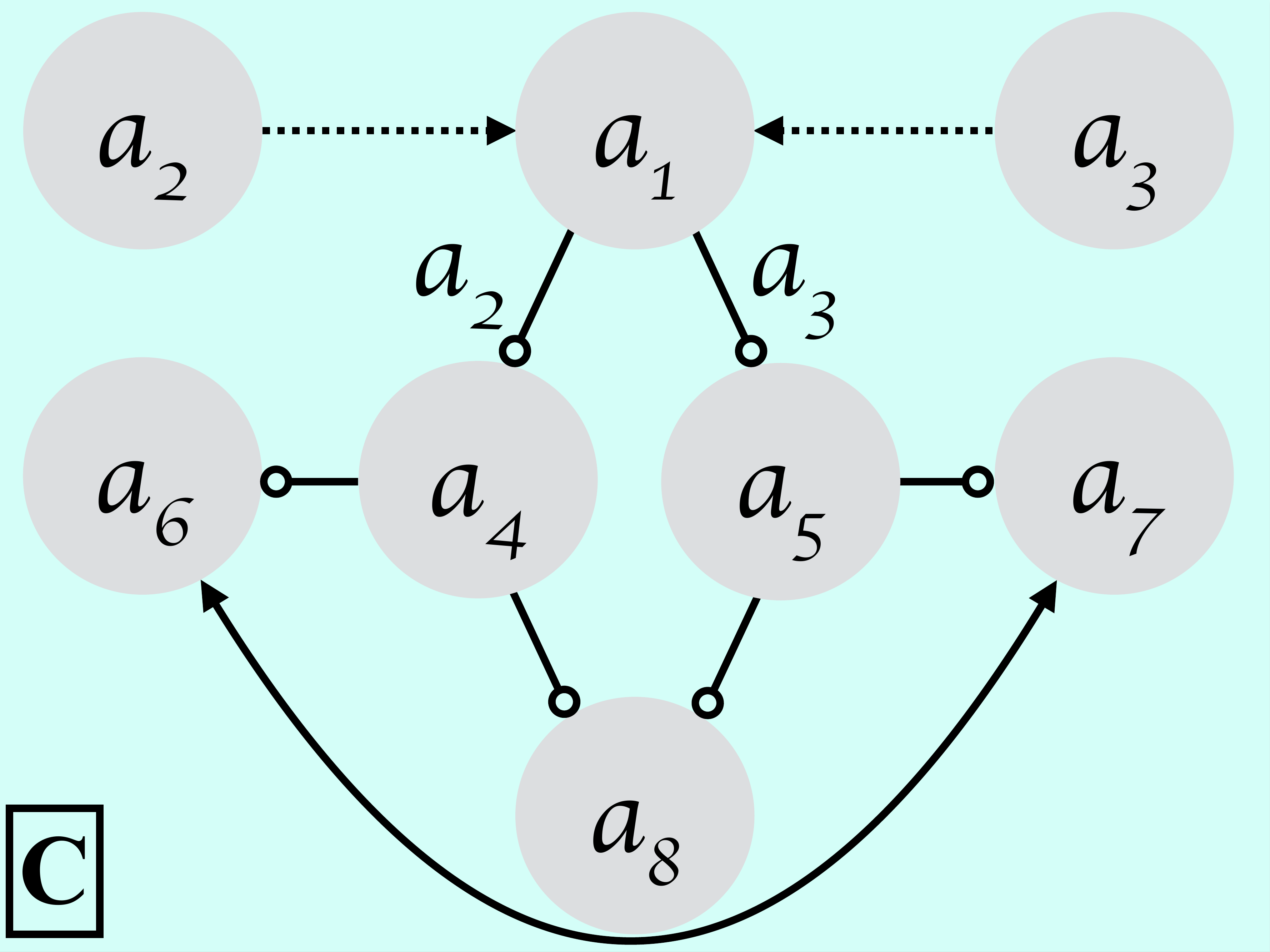}  
\end{center}     
\subsubsection{Defence and reference set}  
One aspect that has not been shed much light on 
in the literature of dynamic argumentation 
is defence against such persuasive acts (dynamic operations).  
Let us consider an example. 
\begin{description} 
   \item[$a_2$] (Mr. X) Elma does not like the music.  
   \item[$a_3$] (Mr. Z) We should get a piano. 
   \item[$a_4$] (Mrs. Y) We can buy Elma a Hello Kitty shoulder bag. 
   \item[$a_5$] (Mrs. Y) We will go to Yamaha Music Communications 
     Co., Ltd. for a piano. 
\end{description}  
The relation among them is as shown in 
Figure \fbox{A} (\fbox{B}): there is an attack 
from $a_2$ to $a_3$, and there is also a persuasion 
act by (Mr. Z holding) $a_3$ trying to 
convert $a_4$ into $a_5$. 
Suppose that $a_5$ is not initially on Mrs. Y's mind,
that is, that it is not visible initially.  
If the persuasion by Mr. Z is successful, 
Mrs. Y changes her mind, dropping $a_4$ and gaining $a_5$. Otherwise, she holds onto 
$a_4$. In Dung theory, 
defence of an argument $a_x$ is defined with respect to 
a set of arguments $A$. The reference set $A$ defends $a_x$ 
just when $A$'s members attack 
all arguments attacking $a_x$. 
We see that this concept may be extended also to persuasion operations. 
For example, if, as marked with a rectangular box in Figure \fbox{A}, the reference set 
consists of $a_4$ and $a_5$ alone, 
it does not detect any flaw in $a_3$. Thus, the persuasion is successful 
with respect to the reference set. 
However, if it also contains $a_2$ 
attacking $a_3$ as in Figure \fbox{B}, it can prevent the persuasion 
from taking effect on $a_4$.  
\subsubsection{Multiple persuasions} We have a kind of concurrency scenario  
when multiple persuasions act on 
an argument. Let us consider an example.   
\begin{description} 
    \item[$a_1$] 
(Alice at London Bridge, having agreed 
to see Bob at 7 pm) I am going to 
have dinner with Bob. It is 7 pm now. He should be arriving soon. 
\item[$a_2$] 
    (Tom, calling from Camden) Chris (Alice's brother) is looking for you. 
He is at Camden Bar. He says there is some urgent matter, can you please get to the bar as soon as possible? 
\item[$a_3$] (Katie, seeing Alice by chance) Hey Alice, 
you've left your laptop at King's library? You better 
go there now. Oh, and don't forget about your presentation  
tomorrow morning. Make sure you have slides\linebreak ready! 
\end{description}  
Having been acquainted with Bob only recently, Alice 
is more inclined to getting to Camden Bar 
or to King's library. That is, $a_4$: I am going to Camden Bar, 
or $a_5$: I am going to King's library. 
She knows her brother is very stern. 
But the assignment of which Katie reminded Alice seems 
to be a thing that must be prioritised, too. Whichever option she is to
go for, she must, thinks she, come up with excuses to justify 
her choice. Therefore: 
\begin{description} 
    \item[$a_6$] (Alice's excuse) 
        It is fine to skip dinner because 
        I waited for Bob at London Bridge and he did not 
        arrive in time. Besides, I suddenly have something urgent. 
    \item[$a_7$] (Alice's excuse) I cannot see Chris. 
        For my career, it is important that I perform 
        well at presentation tomorrow. Chris will 
        understand.  
    \item[$a_8$] (Alice's excuse) I cannot go to King's 
        library now, because it is 
        always urgent when Chris calls me. \\
\end{description}  
Figure \fbox{C} represents these arguments. Now, 
what we have is a potentially irreversible branching. If 
$a_2$ persuades $a_1$ into $a_4$, it is no longer possible 
for $a_3$ to persuade $a_1$, as $a_1$ will not be available 
for persuasion. If $a_3$ persuades $a_1$ into $a_5$, on the other 
hand, it is no longer possible that $a_2$ persuades $a_1$. A 
certain partial order may be defined among persuasion (as in preference-based argumentation), but the non-deterministic consideration leads to a more general theory (as in probabilistic argumentation) 
in which the actual behaviour of a system depends on 
run-time executions. \\
\indent Just 
as in program analysis, however, it may be still possible 
to identify certain properties, whichever an actual path 
may be. In this particular example, 
(Alice holding) $a_1$ may be persuaded into holding 
$a_4$ or else $a_5$, and we cannot tell which with certainty. However, 
we can certainly predict $a_8$'s emergence. Thus, 
by obtaining varieties in arguments admissibility 
by means of CTL, 
we can answer such a query as `Is $a_8$ going to be an admissible 
argument in whatever order persuasive acts may take place?'. 
\section{Technical Backgrounds}   
Let $\mathcal{A}$ be a class of abstract entities which 
we understand as arguments.  We denote any member of $\mathcal{A}$ 
by $a$ with or without a subscript, 
and any finite subset of $\mathcal{A}$ 
by $A$ with or without a subscript. 
An argumentation framework \cite{Dung95} is 
a tuple $(A, R)$ where $R$ is
a binary relation over $A$.  
Let $F^{(A, R)}(A_1)$ 
denote $(A_1, R \cap (A_1 \times A_1))$, 
we denote 
by 
$2^{(A, R)}$ the following set: 
$\bigcup_{A_1 \subseteq A}F^{(A, R)}(A_1)$, i.e. 
all sub-argumentation frameworks 
of $(A, R)$. When confusion is unlikely to occur,  
we abbreviate $F^{(A, R)}(A_1)$ for 
some $A_1$ by $F(A_1)$. \\
\indent For any $(A, R)$ 
an argument $a_1 \in A$
is said to attack $a_2 \in A$ 
if and only if, or iff, $(a_1, a_2) \in R$.     
A set of arguments $A_1 \subseteq A$ is said to 
 defend $a_x \in A$ iff 
each $a_y \in A$ attacking  
$a_x$ is attacked by at least one argument  
in $A_1$. A set of arguments $A_1 \subseteq A$ is said to be: 
    conflict-free iff   
        no member of $A_1$ attacks a member of $A_1$; 
    admissible iff 
        it is conflict-free and it defends  
        all the members of $A_1$; 
 complete iff it is admissible and  
        includes any argument it defends; 
     preferred iff   
        it is a set-theoretically maximal 
        admissible set; 
     stable iff it is preferred and attacks 
        every argument in $A \backslash A_1$; 
    and grounded iff it is the set intersection 
        of all complete sets of $A$.  
\section{Abstract Persuasion Argumentation}     
We define our Abstract Persuasion Argumentation (APA) 
framework 
to be a tuple $(A, R, \Rp, A_0, \hookrightarrow)$ 
for $A_0 \subseteq A$, for a ternary 
relation\linebreak $\Rp:  
 A \times (A \cup \{\epsilon\}) \times 
A$ and for another 
 $\hookrightarrow: 2^A \times (2^{(A, R)} \times 2^{(A, R)})$.
For $\Rp$, 
 $(a_1, \epsilon, a_2) \in \Rp$ 
 represents 
  \mbox{$a_1 \multimap a_2$} 
  (passive persuasion or to induce),  
and  $(a_1, a_2, a_3) \in \Rp$ represents 
\mbox{$a_1 \dashrightarrow a_2 
\lol{a_1} a_3$} (active persuasion or to convert). 
We refer to a subset of $\Rp$ by $\Gamma$ 
with or without a subscript and/or a superscript. \\
\indent APA is a dynamic argumentation framework 
where arguments can appear (go visible) or disappear (go invisible). 
As in a transition system,  
it comes with an initial state and a transition 
relation $\hookrightarrow$.  
For any APA framework $(A, R, \Rp, A_0, \hookrightarrow)$, 
we define a state to be a member $F(A_x)$ of $2^{(A, R)}$, 
and we say any argument that occurs in a state 
{\it visible} and 
any that does not occur in the state 
{\it invisible}, in each case at that particular state.\footnote{We
assume the standard notion of occurrence.}  
We define $F(A_0)$ to be the {\it initial state}. 
\begin{example} 
In Elma example, we assumed 
$A_0 = \{a_2, a_3, a_4\}$ and  \linebreak
$F(A_0) = (A_0, \{(a_2, a_3)\})$. 
In Alice example, 
$A_0 = \{a_1, a_2, a_3\}$ and $F(A_0) = (A_0, \emptyset)$. 
\end{example}  
\begin{definition}[Reachable states]   
For APA $(A, R, \Rp, A_0, \hookrightarrow)$, 
for a set of arguments $A_x \subseteq A$, and for states $F(A_1)$ and $F(A_2)$,
we say that there is a transition from $F(A_1)$ to $F(A_2)$ 
with respect to $A_x$ iff it holds that  
$(A_x, (F(A_1), F(A_2))) \in \hookrightarrow$, 
which we alternatively state either as 
$(F(A_1), F(A_2)) \in \hookrightarrow^{A_x}$
or as $F(A_1) \hookrightarrow^{A_x} F(A_2)$.  
We say that a state $F(A_x)$ is {\it reachable}
iff $F(A_x)$  either is the initial state 
or else is such that 
$F(A_0) \hookrightarrow^{A_{i_1}} 
\cdots \hookrightarrow^{A_{i_n}} F(A_x)$, $1 \leq n$. 
\end{definition}   
A reachable state is a static snapshot of 
an APA framework at one moment, which 
is a Dung argumentation framework. 
To enumerate all reachable states, 
it suffices to define $\hookrightarrow$ 
in specific detail. 
And this is where the notion of defence against 
persuasive acts with 
respect to a reference set at a state - specifically 
visible arguments of the set at the state - comes into play: 
\begin{definition}[Possible persuasion acts]  
 For APA \mbox{$(A, R, \Rp, A_0, \hookrightarrow)$}, 
 we say that a persuasion act $(a_1, \alpha, a_2) \in 
\Rp$, $\alpha \in \{\epsilon\} \cup A$, 
is possible with respect to: (i)  
a reference set $A_x \subseteq A$; and 
(ii) a state $F(A_u)$ 
iff $a_1, \alpha \in A_u \cup \{\epsilon\}$ $\andC$ 
$a_1$ is not attacked by any member of $A_x \cap A_u$.     
We denote the set of all members of 
$\Rp$ that are possible with respect to 
a reference set $A_x \subseteq A$ and 
a state $F(A_u)$ by $\Gamma^{A_x}_{F(A_u)}$. 
\end{definition} 
\begin{example}{\it (Continued)} 
    In Elma example with $A_0 = \{a_2, a_3, a_4\}$,  
   there is one argument, $a_2$, which is in $A_0$ (thus visible), 
   and which attacks $a_3$, 
   so 
  $(a_2, a_3, a_4) \in \Rp$ is possible with respect 
   to $A_x \subseteq A$ and $F(A_0)$ only if 
   $a_2 \not\in A_x$. 
   $\Gamma^{A_x}_{F(A_0)}$ is: $\{(a_2, a_3, a_4)\}$
   if $a_2 \not\in A_x$; $\emptyset$, otherwise.  
  In Alice example with $A_0 = \{a_1, a_2, a_3\}$,
   $(a_2, a_1, a_4)$ and 
   $(a_3, a_1, a_5)$ are both possible
   with respect to any $A_x \subseteq A$ and
    $F(A_0)$, because for no $(a_x, \alpha, a_y) \in \Rp$
   there is $(a_z, a_x) \in R$. 
  $\Gamma^{A_x}_{F(A_0)} = \{(a_2, a_1, a_4), 
   (a_3, a_1, a_5)\}$. 
\end{example}    
Since transition as we consider is non-deterministic,  
each persuasion act possible in a state may or may not 
execute for transition. Therefore, for any 
APA $(A, R, \Rp, A_0, \hookrightarrow)$, any reference 
set $A_x \subseteq A$ 
and any state $F(A_1)$, there are $2^{|\Gamma^{A_x}_{F(A_1)}|} - 1$  
transitions, though some of them may be identical.  
\begin{definition}[Non-deterministic transition] {\ }\\
For APA $(A, R, \Rp, A_0, \hookrightarrow)$, 
for $A_1 \subseteq A$  
and for $\Gamma \subseteq \Rp$,  
let $\nve^{A_1}(\Gamma)$ 
   be $\{a_x \in A_1 \ | \ \exists a_1, a_2 \in A_1.(a_1, a_x, a_2) 
    \in \Gamma\}$, 
   and let $\pve^{A_1}(\Gamma)$  
   be $\{a_2 \in A_1\ | \ \exists a_1, \alpha \in A_1 \cup 
   \{\epsilon\}.(a_1, \alpha, a_2)
   \in \Gamma\}$. 
   For $A_x \subseteq A$ and states $F(A_1)$ and $F(A_2)$, we define: 
   $F(A_1) \hookrightarrow^{A_x} F(A_2)$  iff
   there is some $\emptyset \subset \Gamma \subseteq 
\Gamma^{A_x}_{F(A_1)} \subseteq \Rp$
   such that 
   $A_2 = (A_1 \backslash \nve^{A_1}(\Gamma))
    \cup \pve^{A_1}(\Gamma)$. 
\end{definition}    
For $\Gamma \subseteq \Rp$, if $\Gamma \subseteq \Gamma^{A_x}_{F(A_1)}$, 
it is a (non-deterministically) chosen set of 
possible persuasion acts at $F(A_1)$. 
Thus, $\nve^{A_1}(\Gamma)$ is the set of 
all visible arguments that are to be converted, 
and $\pve^{A_1}(\Gamma)$ is that of all visible 
arguments that are to be generated, in the transition. 
As clear from this definition, 
while every member of $\pve^{A_1}(\Gamma)$, 
if not visible in $F(A_1)$, will be visible in $F(A_2)$,  
not necessarily every member of $\nve^{A_1}(\Gamma)$ 
will be invisible in $F(A_2)$ in case 
it also belongs to $\pve^{A_1}(\Gamma)$, in which case 
the effect is offset. 
\begin{example} {\ }
Consider the argumentation 
in the figures below, in each of which 
visible arguments are marked with a black border around 
the circle. 
Suppose $A_0 = \{a_1, a_2\}$ as in Figure \fbox{D}. 
At $F(A_0)$, there are more than one possible 
persuasion acts: 
$\Gamma^{A_x}_{F(A_0)} = \{(a_1, a_2, a_4), 
(a_2, a_1, a_3)\}$ for any reference set $A_x \subseteq A$. 
There are three transitions for $F(A_0)$, depending 
on which one(s) execute simultaneously. 
If just $(a_1, a_2, a_4)$, $a_2$ will go invisible, while 
$a_4$ will be visible, so we have $F(A_0) \hookrightarrow^{A_x} 
F(A_1)$ (Figure \fbox{E}) for any $A_x$. If just $(a_2, a_1, a_3)$, 
we have $F(A_0) \hookrightarrow^{A_y} F(A_2)$ (Figure 
\fbox{F}) for any $A_y$. Or both of them may execute at once, 
in which case both $a_1$ and $a_2$ will be invisible, 
and $a_3$ and $a_4$ meanwhile will be visible, 
so we have $F(A_0) \hookrightarrow^{A_z} F(A_3)$ (Figure 
\fbox{G}) for any $A_z$. Reasoning similarly 
for the new states, we eventually enumerate 
all reachable states and all transitions among them:  
   \begin{center} 
    \includegraphics[scale=0.11]{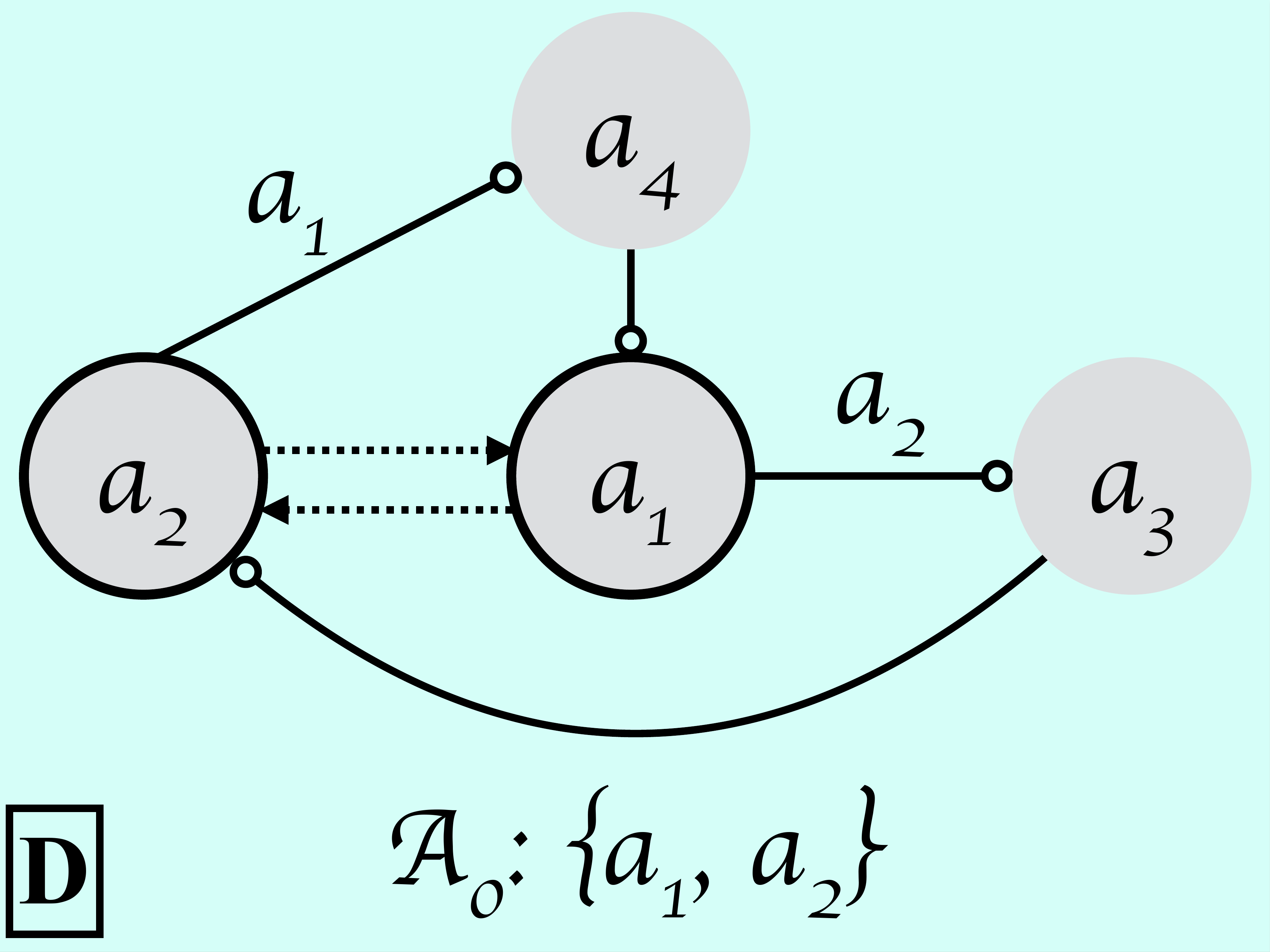}  
    \includegraphics[scale=0.11]{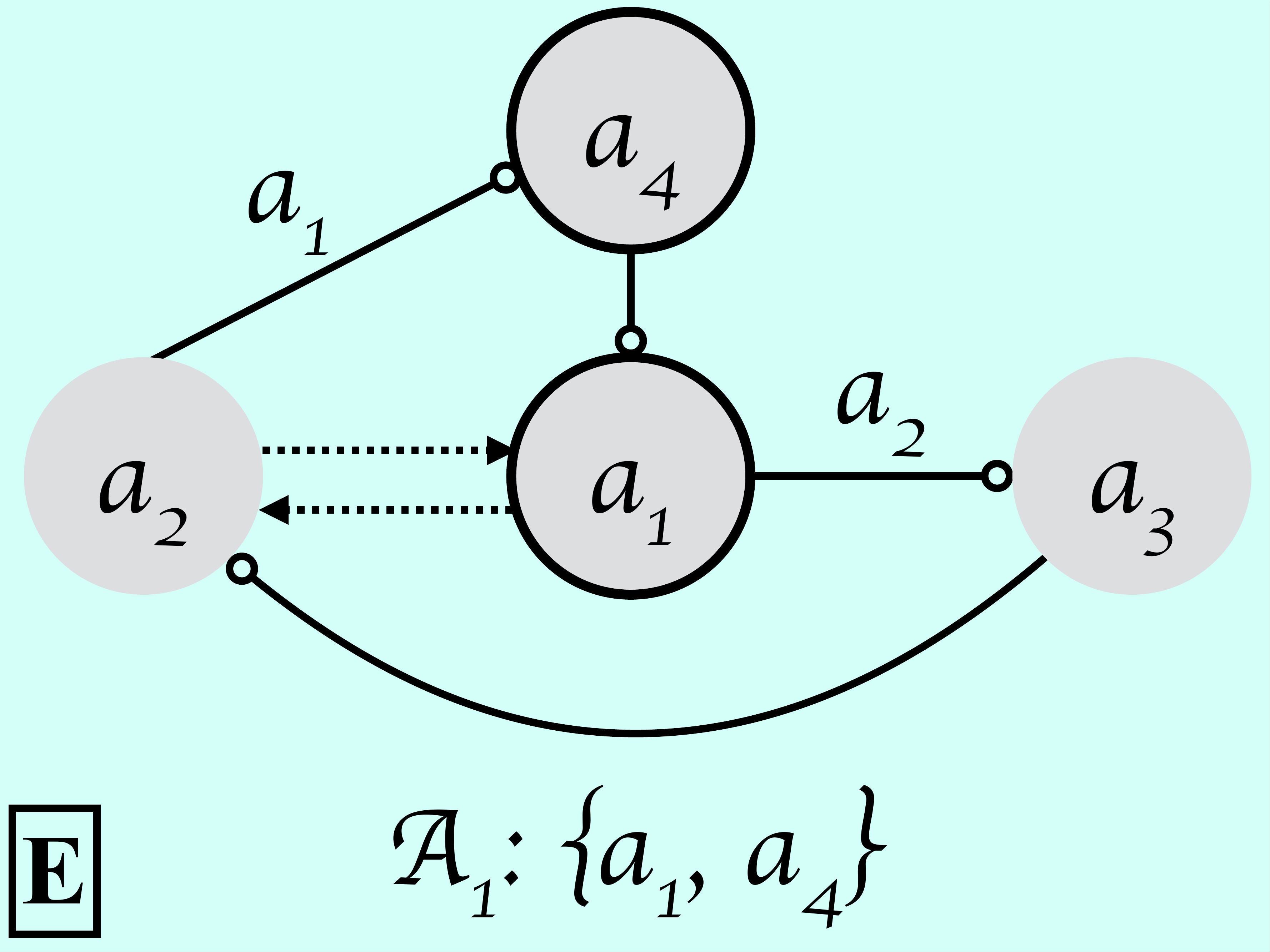}
    \includegraphics[scale=0.11]{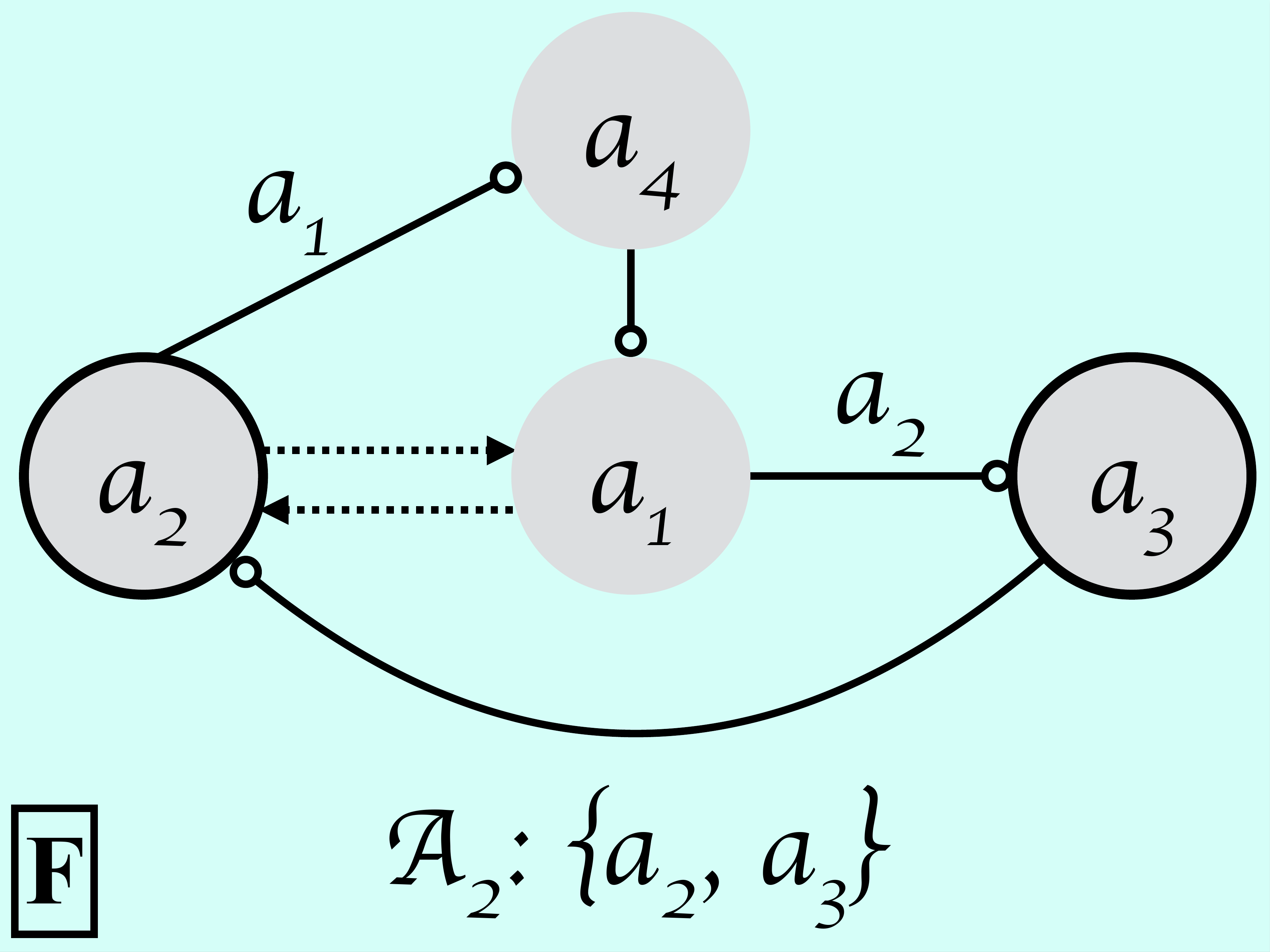} 
    \includegraphics[scale=0.11]{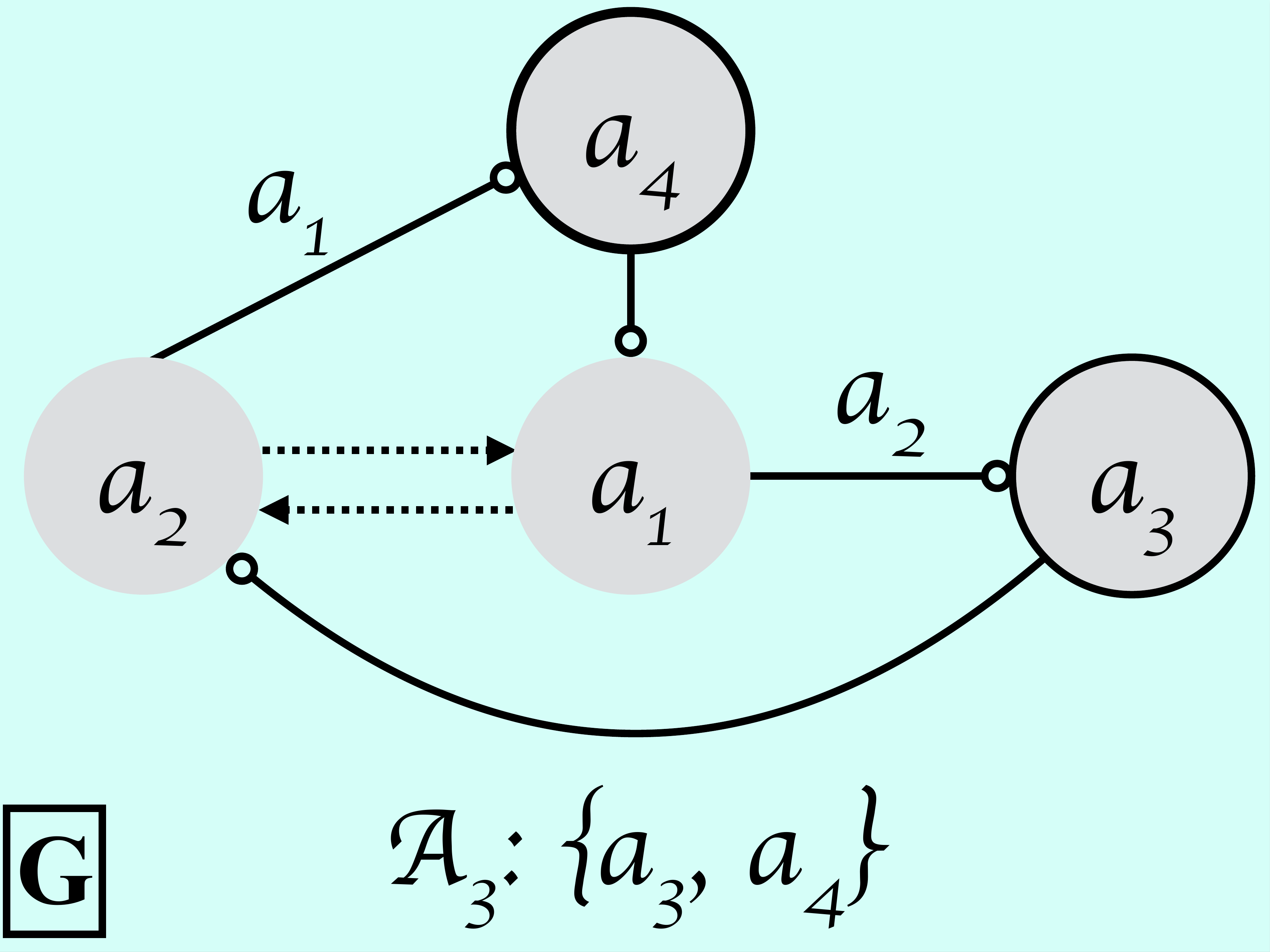}
    \includegraphics[scale=0.11]{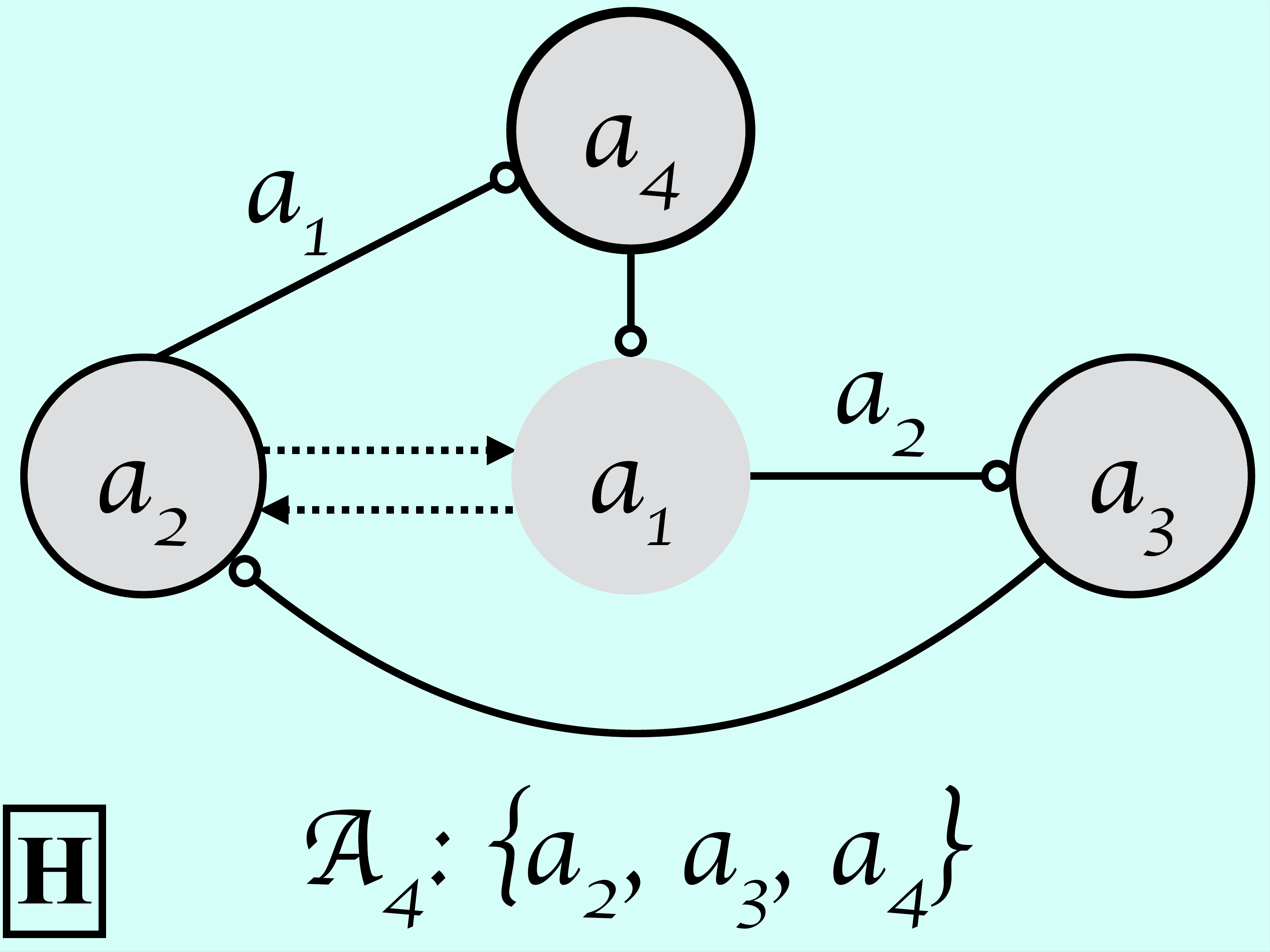}  
    \includegraphics[scale=0.11]{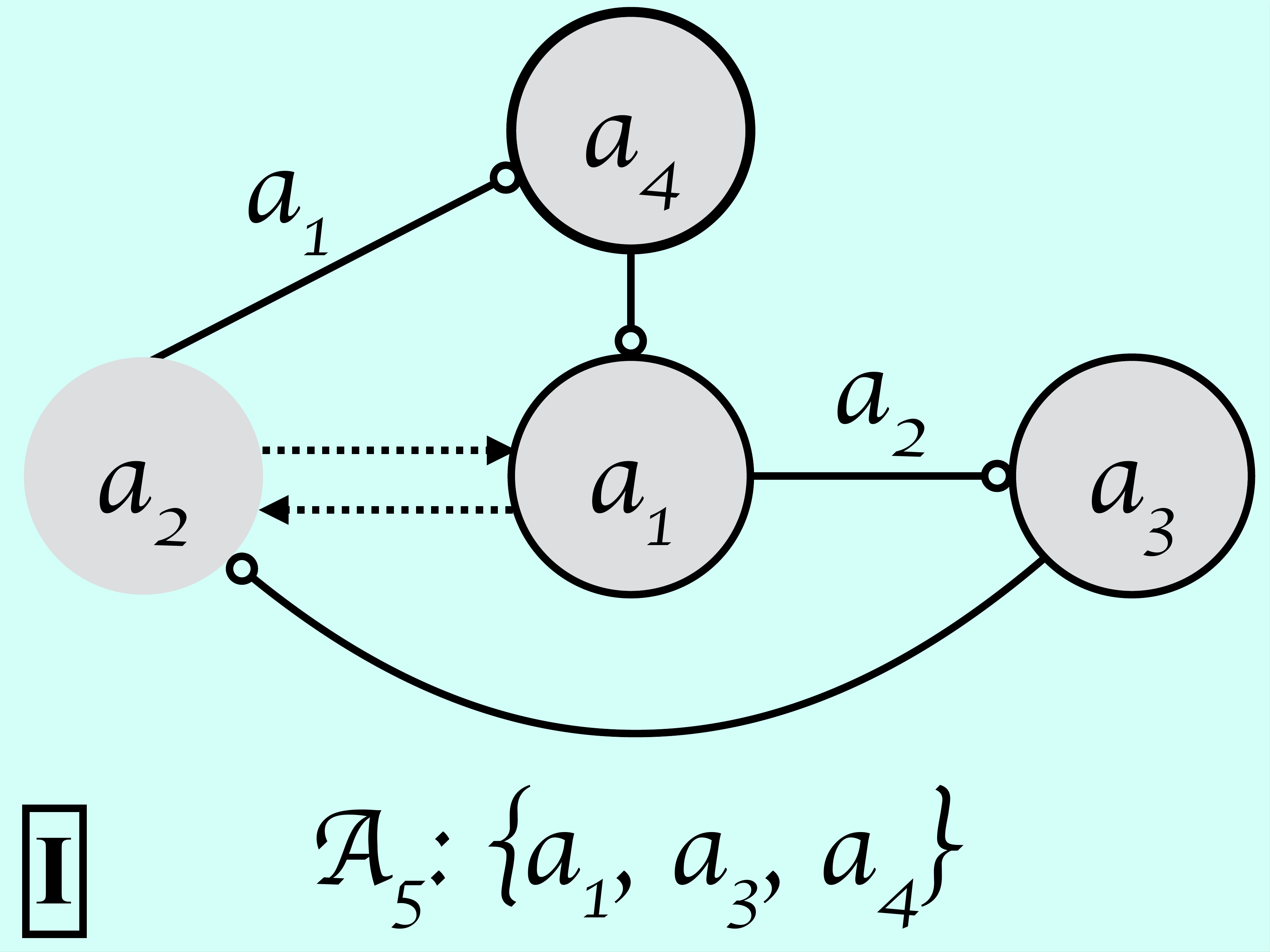}
    \includegraphics[scale=0.11]{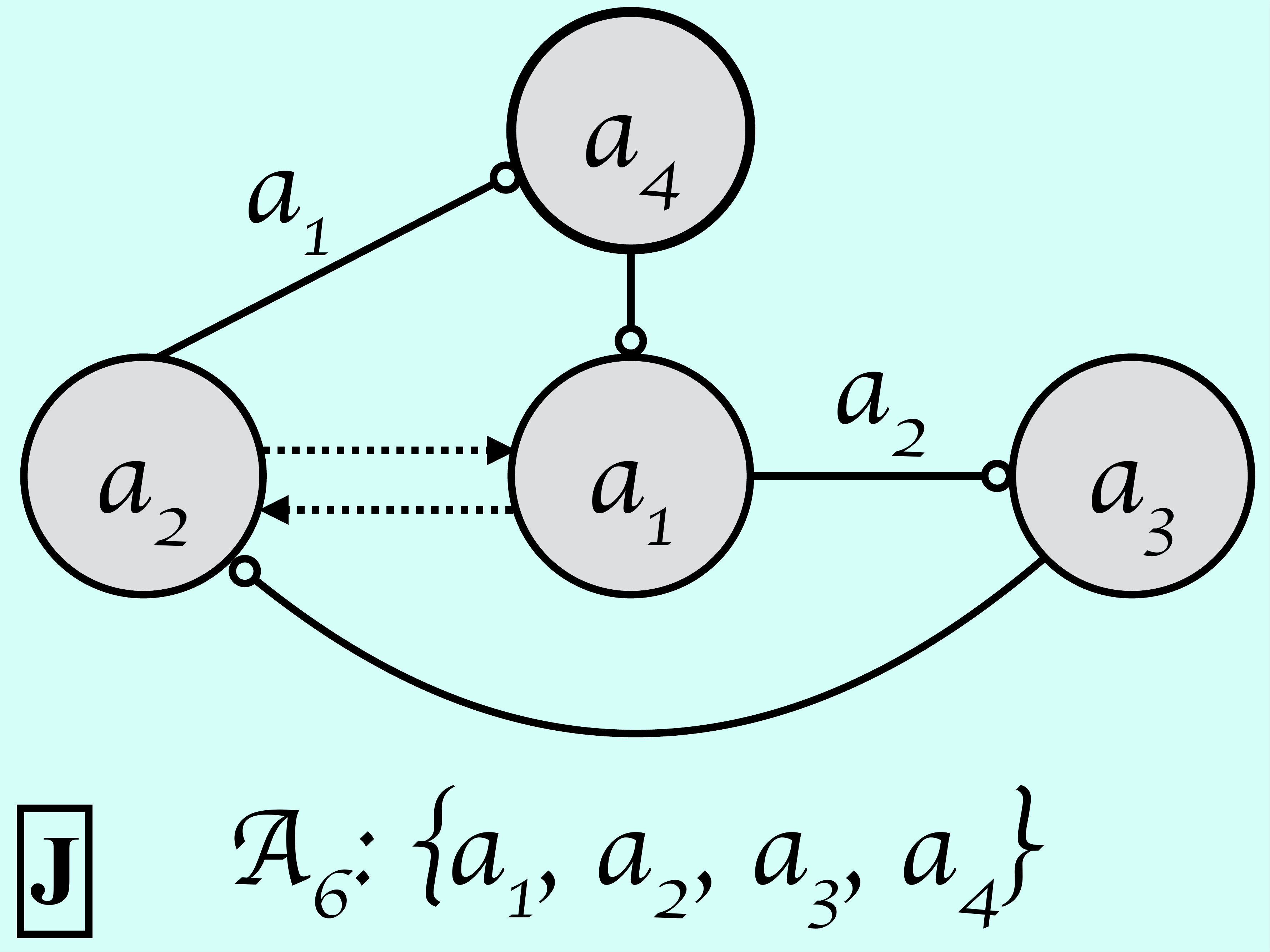} 
    \includegraphics[scale=0.11]{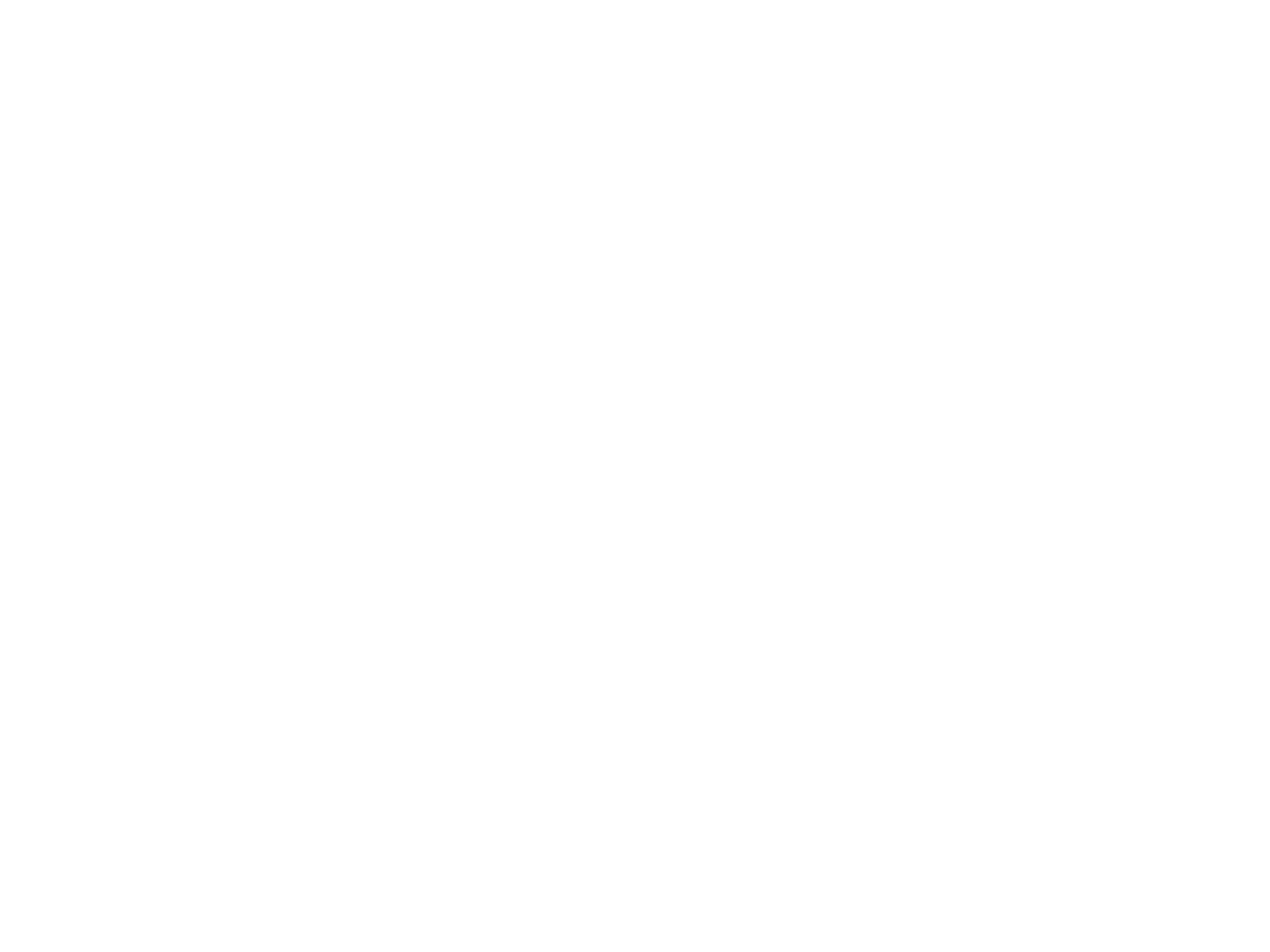}
    \includegraphics[scale=0.11]{blank.pdf} 
\end{center}     
\begin{itemize} 
   \item $F(A_0) \hookrightarrow^{A_x} F(A_1)$, 
      $F(A_0) \hookrightarrow^{A_y} F(A_2)$, 
      $F(A_0) \hookrightarrow^{A_z} F(A_3)$.
   \item $F(A_1) \hookrightarrow^{A_u} F(A_1)$.
   \item $F(A_2) \hookrightarrow^{A_v} F(A_2)$.
   \item $F(A_3) \hookrightarrow^{A_p} F(A_4)$, 
      $F(A_3) \hookrightarrow^{A_q} F(A_5)$, 
$F(A_3) \hookrightarrow^{A_r} F(A_6)$.
   \item $F(A_4) \hookrightarrow^{A_i} F(A_4)$,  
$F(A_4) \hookrightarrow^{A_j} F(A_6)$.
   \item $F(A_5) \hookrightarrow^{A_k} F(A_5)$, 
       $F(A_5) \hookrightarrow^{A_c} F(A_6)$.
   \item $F(A_6) \hookrightarrow^{A_d} F(A_4)$, 
 $F(A_6) \hookrightarrow^{A_f} F(A_5)$, 
$F(A_6) \hookrightarrow^{A_g} F(A_6)$.  
\end{itemize}  
The reference sets for the transitions are 
any subset of $\{a_1, a_2, a_3, a_4\}$.  
Notice, apart from the trivial self-transitions, 
states could oscillate infinitely between 
$F(A_4), F(A_5)$ and $F(A_6)$. 
\end{example} 
\begin{proposition}
    Suppose an APA framework $\delta$ with a finite number of arguments. 
    It is necessary that 
    the number of (reachable) states is finite. 
It is, however,  not necessary that the number of transitions in $\delta$ 
    is finite. \end{proposition}  
\subsection{Admissibilities}    
We now define the static notion of admissibility in APA frameworks, 
based on three criteria. For APA $(A, R, \Rp, A_0, \hookrightarrow)$, 
\\\\
\textbf{Conflict-freeness}  
We say that $A_1 \subseteq A$ is conflict-free 
in a (reachable) state $F(A_x)$ iff 
no member of $A_1 \cap A_x$ attacks 
a member of $A_1 \cap A_x$. \\\\
\textbf{Defendedness} 
We say that a reference set $A_1 \subseteq A$ defends $a \in A$
in a state $F(A_x)$ iff either $a \not\in A_x$ or else both of 
the conditions below hold. 
\begin{enumerate}
\item Every $a_u \in A_x$ attacking $a$ is attacked 
by at least one member of $A_x \cap A_1$ (counter-attack). 
\item There is no state $F(A_y)$ 
such that both $F(A_x) \hookrightarrow^{A_1} F(A_y)$ 
and $a \not\in A_y$ at once (no elimination). 
\end{enumerate} 
We say that $A_1 \subseteq A$ is defended in a state $F(A_x)$ iff 
$A_1$ as a reference set defends every 
member of its in $F(A_x)$. \\\\
\textbf{Properness} 
We say that  $A_u \subseteq A$ is proper 
in a state $F(A_x)$ 
iff $A_u \subseteq A_x$.  \\\\
Defendedness above  
extends Dung's defendedness naturally for 
$\Rp$. \linebreak
Properness ensures that we will not be talking of invisible 
arguments. 
With these properties, we say 
$A_u \subseteq A$ is: admissible in a state $F(A_1)$ iff 
$A_u$ is conflict-free, defended and 
proper in $F(A_1)$; complete in a state $F(A_1)$ 
iff 
$A_u$ is admissible in $F(A_1)$ and includes all arguments  
it defends; preferred iff 
no $A_v \subseteq A$ that is complete 
in a state $F(A_1)$ 
is a strict superset of $A_u$; stable 
iff it is preferred and attacks every member of 
$A_1 \backslash A_u$; 
and grounded in a state $F(A_1)$ iff it is the set intersection 
of all complete sets in $F(A_1)$. Since 
each state is a Dung argumentation framework, we have: 
\begin{proposition}  
  For APA $(A, R, \Rp, A_0, \hookrightarrow)$, for 
  a state $F(A_1)$ and for $A_x \subseteq A_1$: 
   if $A_x$ is stable, 
    then $A_x$ is preferred;  
    if $A_x$ is preferred, then 
    $A_x$ is complete; 
    if $A_x$ is complete, then 
    $A_x$ is admissible; there exists 
    at least one complete set; 
    and there may not exist any stable set. 
\end{proposition}   
For general admissibilities across transition, 
one way of describing more varieties
is to embed this state-wise admissibility and transitions 
into computation tree logic
(CTL) or other branching-time logic, by which 
model-theoretical results known to the logic become available to APA 
frameworks, too.  We consider CTL with some path restrictions. 
Denote $\{\textsf{ad}, \textsf{co}, \textsf{pr}, \textsf{st}, 
 \textsf{gr}\}$ by $\Omega$, 
and refer to a member of $\Omega$ by $\omega$.  
Let the grammar of $\phi$ be:  \\
\indent $\phi := \bot \ | \ 
\top \ | \ a \inAPA A_x \ | \ P_\delta(\omega, A_x) \ | \  
\neg \phi \ | \ \phi \wedge \phi \ | \ \phi \vee \phi \ | \ 
\phi \supset \phi \ | \
\textbf{AX}^{\Sigma} \phi \ | \\
{\ }\quad\quad 
\textbf{EX}^{\Sigma}
 \phi \ | \ 
\textbf{AF}^{\Sigma} \phi \ | \ \textbf{EF}^{\Sigma} \phi  \ | \ 
\textbf{AG}^{\Sigma} \phi \ | \
\textbf{EG}^{\Sigma} \phi \ | \ \textbf{E}^{\Sigma}[\phi\textbf{U}\phi] \ | \ 
\textbf{A}^{\Sigma}[\phi\textbf{U}\phi]$\\\\
where both $a \inAPA A_x$ and 
$P_{\delta}(\omega, A_x)$ are atomic predicates 
for an APA framework $\delta := 
(A, R, \Rp, A_0, \hookrightarrow)$, with $A_x \subseteq A$, 
with $a \in A$ and with $\Sigma \subseteq 2^A$. 
$\textbf{A}$ is `in all branches', 
$\textbf{E}$ is `in some branch', 
$\textbf{X}$ is `next state', 
$\textbf{F}$ is `future state',
$\textbf{G}$ is `all subsequent states', 
and $\textbf{U}$ is `until'. 
The superscripts restrict paths to only those 
reachable with member(s) of $\Sigma$ as reference set(s). See 
below for the exact semantics. 
We denote the class of all atomic predicates 
for $\delta$ by $\mathcal{P}_\delta$.
For semantics, 
let $L: \Omega \times 2^{\mathcal{A}} \rightarrow 2^{\mathcal{P}_\delta}$ be a 
valuation function such that  $L(\omega, A_x)$ is: 
\begin{itemize} 
    \item $\{P_\delta(\textsf{ad}, A_y) \in \mathcal{P}_\delta\ | \ 
    A_y \text{ is admissible in } 
    F(A_x)\}$ if $\omega = \textsf{ad}$. 
    \item $\{P_\delta(\textsf{co}, A_y) \in \mathcal{P}_\delta\ | \ 
    A_y \text{ is complete in } 
    F(A_x)\}$ if $\omega = \textsf{co}$. 
    \item $\{P_\delta(\textsf{pr}, A_y) \in \mathcal{P}_\delta\ | \ 
    A_y \text{ is preferred in } 
    F(A_x)\}$ if $\omega = \textsf{pr}$. 
    \item $\{P_\delta(\textsf{st}, A_y) \in \mathcal{P}_\delta\ | \ 
    A_y \text{ is stable in } 
    F(A_x)\}$ if $\omega = \textsf{st}$. 
    \item $\{P_\delta(\textsf{gr}, A_y) \in \mathcal{P}_\delta\ | \ 
    A_y \text{ is grounded in } 
    F(A_x)\}$ if $\omega = \textsf{gr}$. 
\end{itemize} 
We define $\mathcal{M} := (\delta, L)$ to be a 
transition system with 
the following forcing relations.\footnote{
The liberty of allowing arguments 
into $\mathcal{M}$ causes no confusion, let alone issues. 
If one is so inclined, he/she 
may choose to consider that components of $\delta$  
that appear in $\mathcal{M}$ are semantic counterparts 
of those that appear in the syntax of CTL with
one-to-one correspondence between them.} 
\begin{itemize}[leftmargin=0.2cm] 
    \item $\mathcal{M}, A_1 \models \top$.
    \item $\mathcal{M}, A_1 \not\models \bot$.    
    \item $\mathcal{M}, A_1 \models
    a \inAPA A_x$ iff $a \in A_x$. 
    \item $\mathcal{M}, A_1 \models 
                P_\delta(\omega, A_x)$ 
                iff $P_\delta(\omega, A_x) \in L(\omega, A_1)$ (in plain terms, 
                this says $A_x$ is 
                admissible / complete / preferred / stable / 
   grounded in a state $F(A_1)$). 
    \item $\mathcal{M}, A_1 \models \neg \phi$ 
        iff $\mathcal{M}, A_1 \not\models \phi$.  
    \item $\mathcal{M}, A_1 \models 
   \phi_1 \wedge \phi_2$ 
        iff $\mathcal{M}, A_1 \models \phi_1$ 
        $\andC$ $\mathcal{M}, A_1 \models \phi_2$. 
    \item $\mathcal{M}, A_1 \models \phi_1 \vee \phi_2$ 
        iff $\mathcal{M}, A_1 \models \phi_1$ 
        $\orC$ $\mathcal{M}, A_1 \models \phi_2$. 
    \item $\mathcal{M}, A_1 \models \phi_1 \supset \phi_2$
        iff $\mathcal{M}, A_1 \not\models \phi_1$ 
        $\orC$ $\mathcal{M}, A_1 \models \phi_2$. 
    \item $\mathcal{M}, A_1 \models \textbf{AX}^{\Sigma} \phi$ 
        iff $\mathcal{M}, A_2 \models \phi$ for each 
   transition $F(A_1) \hookrightarrowR{A_x} F(A_2)$, $A_x \in \Sigma$.  
    \item $\mathcal{M}, A_1 \models \textbf{EX}^{\Sigma} \phi$  
        iff there is some transition 
 $F(A_1) \hookrightarrowR{A_x} F(A_2)$, $A_x \in \Sigma$, such that 
  $\mathcal{M}, A_2 \models  
        \phi$. 
      \item $\mathcal{M}, A_1 \models \textbf{AF}^{\Sigma} \phi$ 
        iff there is some $i \geq 0$ for each transition
    $F(A_1) \hookrightarrow^{A_{j1}} \cdots \hookrightarrowR{A_{ji}}
    F(A_{i+1}) (\hookrightarrowR{A_x} \cdots)$,
     $A_{jk} \in \Sigma$ for $1 \leq k \leq {i+1}$, 
    such that $\mathcal{M}, A_{i+1} 
    \models \phi$. 
    \item $\mathcal{M}, A_1 \models \textbf{EF}^{\Sigma} \phi$ 
        iff  there are some $i \geq 1$ and 
    a transition
    $F(A_1) \hookrightarrow^{A_{j1}} \cdots \hookrightarrowR{A_{ji}}
    F(A_{i+1}) (\hookrightarrowR{A_x} \cdots)$, 
    $A_{jk} \in \Sigma$ for $1 \leq k \leq {i+1}$, such 
   that $\mathcal{M}, A_{i+1} 
    \models \phi$. 
    \item $\mathcal{M}, A_1 \models \textbf{AG}^{\Sigma} \phi$
        iff $\mathcal{M}, A_k \models \phi$ for each 
   transition $F(A_1) \hookrightarrowR{A_{j1}} 
   \cdots$, $A_{jn} \in \Sigma$ for $1 \leq n$, such that $F(A_k)$ 
   occurs in the transition sequence. 
    \item $\mathcal{M}, A_1 \models \textbf{EG}^{\Sigma}
     \phi$ 
        iff there is some transition 
   $F(A_1) \hookrightarrowR{A_{j1}} \cdots$, 
     $A_{jn} \in \Sigma$ for $1 \leq n$, 
   such that 
 $\mathcal{M}, A_k \models \phi$ 
   and that $F(A_k)$ occurs in the transition sequence.  
   \item $\mathcal{M}, A_1 \models \textbf{A}^{\Sigma}[\phi_1\textbf{U}
    \phi_2]$ iff there exists some $i \geq 0$ for each 
    transition $F(A_1) \hookrightarrowR{A_{j1}} 
   \cdots \hookrightarrowR{A_{ji}} F(A_{i+1}) (\hookrightarrowR{A_x} 
   \cdots)$ such that $\mathcal{M}, A_{i+1} \models \phi_2$
    and that $\mathcal{M}, A_k \models \phi_1$ for all 
    $k < {i+1}$.  
    \item $\mathcal{M}, A_1 \models \textbf{E}^{\Sigma}[\phi_1\textbf{U}      \phi_2]$ iff there exists some $i \geq 0$ and
    a transition $F(A_1) \hookrightarrowR{A_{j1}} \cdots 
    \hookrightarrowR{A_{ji}} F(A_{i+1}) (\hookrightarrowR{A_x} 
   \cdots)$ such that $\mathcal{M}, A_{i+1} \models \phi_2$
   and that $\mathcal{M}, A_{k} \models \phi_1$
     for all $k < {i+1}$. 
\end{itemize}  
\noindent We say that $\phi$ is true (in $\delta$) iff 
$(\delta, L), A_0 \models \phi$. \\
\indent While this logic appears more graded than 
CTL for the superscripts $\Sigma$, there is an obvious encoding of it 
into the standard CTL with an additional atomic predicate in the grammar of $\phi$ 
that judges whether an argument is visible. That is, 
we can for example replace $\textbf{EX}^{\{A_x\}}\phi$
with $\textbf{EX}(\phi_1 \wedge \phi) \vee \cdots 
\vee 
\textbf{EX}(\phi_n \wedge \phi)$ if we can express   
by the expression that, for any transition $F(A_c) \hookrightarrowR{A_x} F(A_d)$ 
such that $F(A_c)$ is the state with respect to which the expression
is evaluated, there exists some $1 \leq i \leq n$ 
such that $\phi_i$ holds good just when all and only members of $A_d$ 
are visible, and that for every $\phi_i$, $1 \leq i \leq n$, 
there exists some $A_d$ such that 
$F(A_c) \hookrightarrowR{A_x} F(A_d)$ and that 
$\phi_i$ holds good just when all members of $A_d$ 
are visible, which confirms 
that our logic is   
effectively CTL. It is straightforward to see the following 
well-known equivalences in our semantics: 
\begin{proposition}[De Morgan's Laws and Expansion Laws] {\ }\\ 
$\neg \textbf{AF}^{\Sigma} \phi \equiv 
\textbf{EG}^{\Sigma} \neg \phi$, $\neg \textbf{EF}^{\Sigma}
 \phi \equiv \textbf{AG}^{\Sigma} \neg \phi$, $\neg \textbf{AX}^{\Sigma}
 \phi \equiv 
\textbf{EG}^{\Sigma} \neg \phi$ (De Morgan's Laws),  
$\textbf{AG}^{\Sigma} \phi \equiv \phi \wedge \textbf{AX}^{\Sigma}\textbf{AG}^{\Sigma} \phi$, 
$\textbf{EG}^{\Sigma} \phi \equiv \phi \wedge \textbf{EX}^{\Sigma}\textbf{EG}^{\Sigma} \phi$, 
$\textbf{AF}^{\Sigma} \phi \equiv \phi \vee \textbf{AX}^{\Sigma}\textbf{AF}^{\Sigma} \phi$, 
$\textbf{EF}^{\Sigma} \phi \equiv \phi \vee \textbf{EX}^{\Sigma}\textbf{EF}^{\Sigma} \phi$,
$\textbf{A}^{\Sigma}[\phi_1 \textbf{U}\phi_2] \equiv 
\phi_2 \vee (\phi_1 \wedge \textbf{AX}^{\Sigma}\textbf{A}^{\Sigma}[\phi_1\textbf{U}\phi_2])$, 
$\textbf{E}^{\Sigma}[\phi_1 \textbf{U}\phi_2] \equiv 
\phi_2 \vee (\phi_1 \wedge \textbf{EX}^{\Sigma}\textbf{E}^{\Sigma}
[\phi_1\textbf{U}\phi_2])$
(Expansion Laws) . 
\end{proposition}  
Proof is by induction on the size (the number of 
symbols) of $\phi$ for each $\Sigma$.  
Other well-known general properties of CTL 
immediately hold true, 
such as existence of a sound and complete axiomatisation of CTL. 
Atomic entailments are decidable for any 
APA $\delta$ (with a finite number of arguments), 
since each state is a Dung argumentation framework. 
\begin{center}
  \includegraphics[scale=0.11]{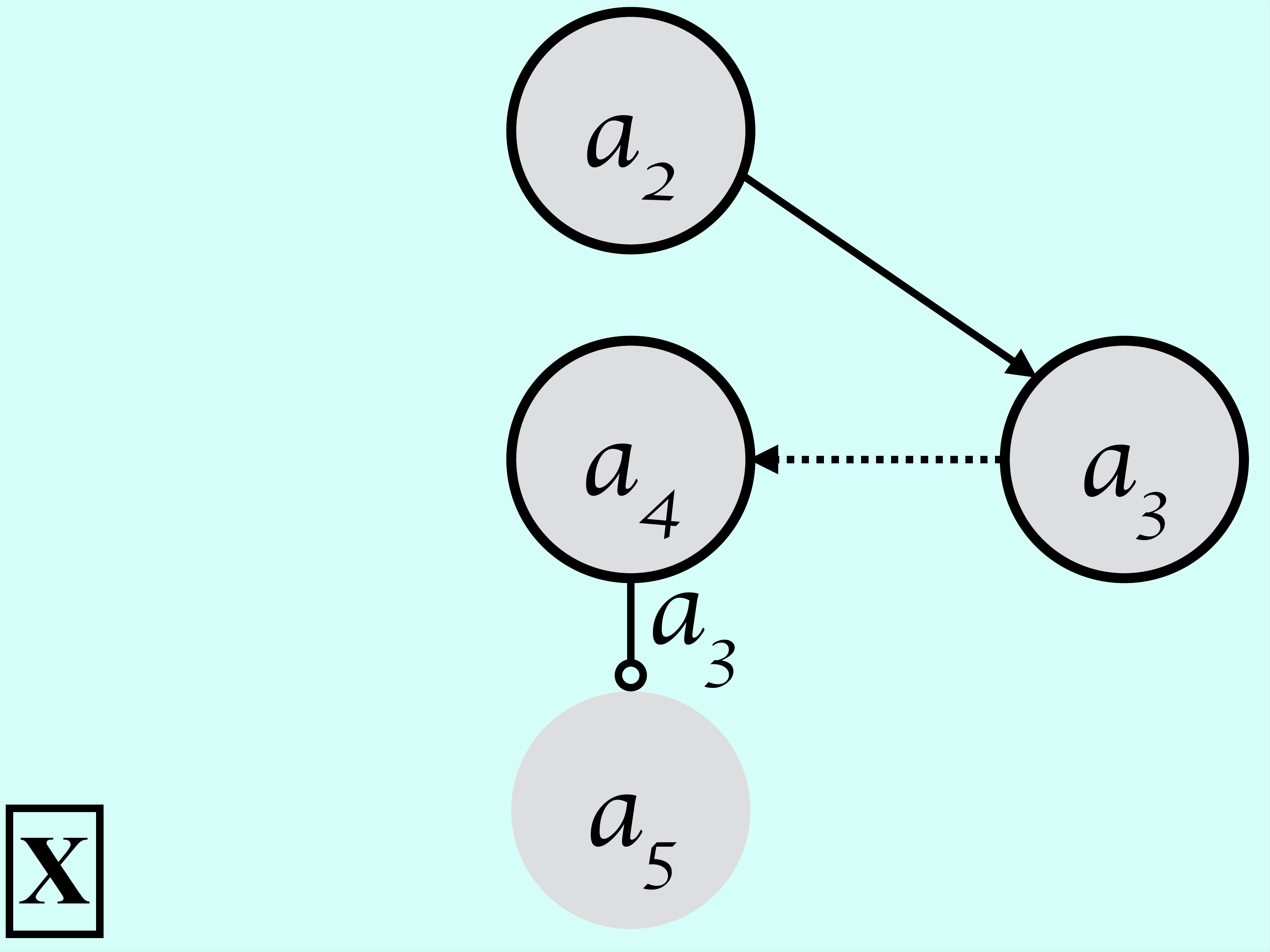}
  \includegraphics[scale=0.11]{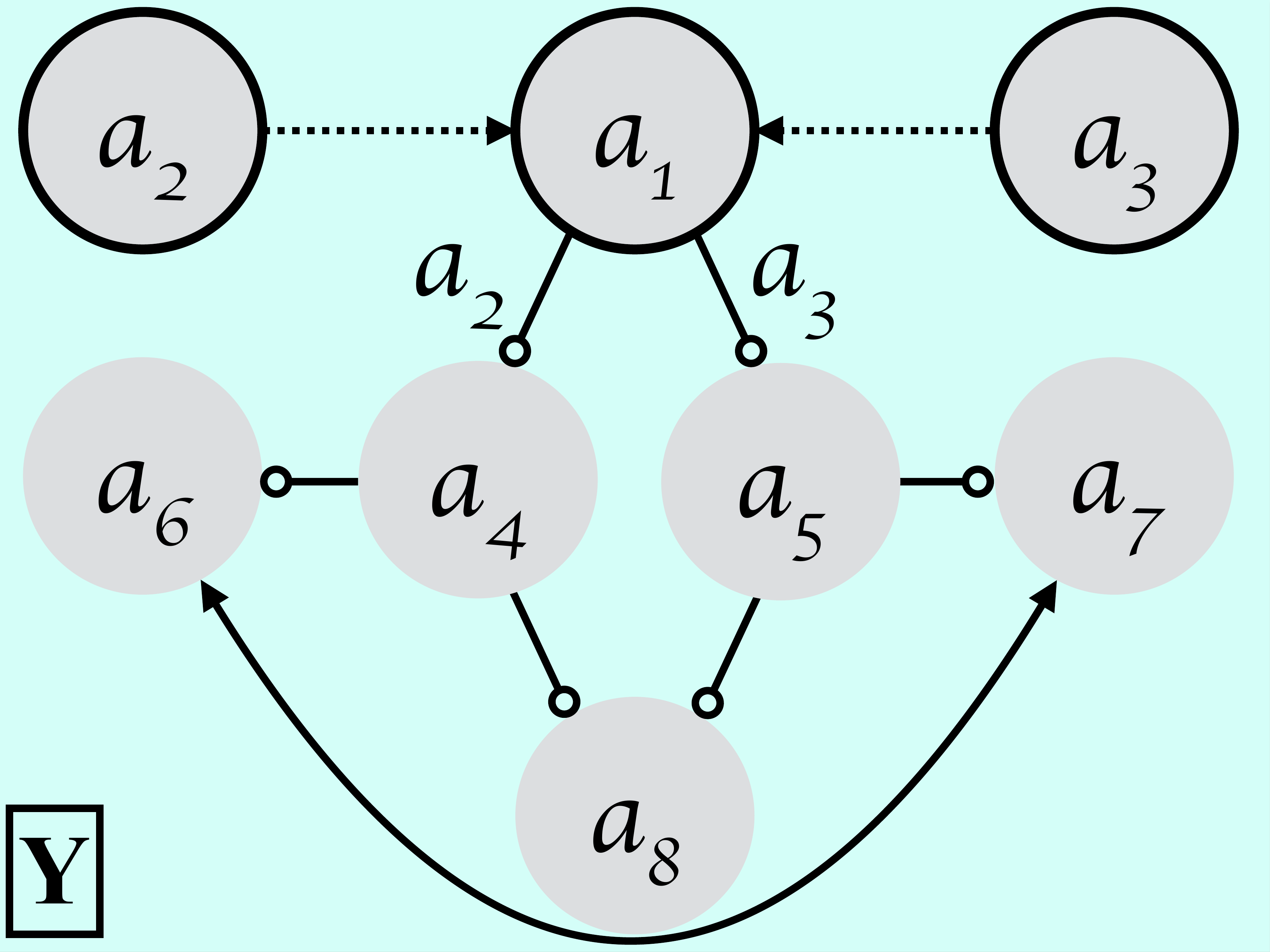}
  \includegraphics[scale=0.11]{figure110.pdf} 
\end{center}
\begin{example} 
    For Elma example (re-listed above in Figure \fbox{X}
    that marks initially visible arguments), recall 
   $A_0 = \{a_2, a_3, a_4\}$. Denote  
    the argumentation by 
    $\delta$.
    By   
    stating that $(a_5 \inAPA A_1 \wedge \textbf{EF}^{\{A_1\}} P_{\delta}(\textsf{ad}, A_1)) \supset 
     \neg a_2 \inAPA A_1$ is true, 
    we have stated that  
    if $a_2$ is a member of a reference set $A_1$, 
    and if the same reference set is used for all transitions,  
    $A_1$ that contains $a_5$ is never admissible.  \\
    \indent For Alice example (re-listed above in
    Figure \fbox{Y} that marks initially visible arguments), recall $A_0 = \{a_1, a_2, a_3\}$. 
    Denote the argumentation by $\delta$. By stating
    that $(\textbf{AF}^{\{A_1\}} \textbf{AG}^{\{A_1\}}P_{\delta}(\textsf{co}, A_1)) \supset
      a_8 \inAPA A_1$ is true, we have stated that  
     if a set of arguments $A_1$ is such that, 
   in all branches with $A_1$ as the reference set,  
   it will be permanently complete 
     from some state on, then it must include $a_8$. \\
\indent For the example in Figure \fbox{D} (re-listed above),   
    recall $A_0 = \{a_1, a_2\}$ within $A = \{a_1, a_2, a_3, a_4\}$. 
    Assume:
   $\exact(\{a_{i_1}, \ldots, a_{i_n}\}, A_2) \equiv \linebreak
    ((a_{i_1} \inAPA A_2) \wedge \cdots \wedge 
    (a_{i_n} \inAPA A_2)) \wedge 
    (\neg a_{i_{n+1}} \inAPA A_2 \wedge \cdots \wedge 
     \neg a_{i_{m}} \inAPA A_2)$ 
     where $A_2 \subseteq A$ and where 
     $\{a_{i_1}, \ldots, a_{i_m}\} \equiv \{a_1, \ldots, a_4\} \equiv A$.
 Assume also that $\Sigma = 2^A$. \\
   \indent 
 By 
     $\exact(\{a_2, a_3, a_4\}, A_4) 
  \wedge \exact(\{a_1, a_3, a_4\}, A_5) \wedge \\
   \indent\indent \textbf{AG}^{\Sigma}((P_{\delta}(\textsf{ad}, A_4)
      \supset \textbf{EF}^{\Sigma}(P_{\delta}(\textsf{ad}, A_5)
 \wedge \neg P_{\delta}(\textsf{ad}, A_4)))\\   
     \indent\indent\indent \wedge (P_{\delta}(\textsf{ad}, A_5)
     \supset \textbf{EF}^{\Sigma}(P_{\delta}(\textsf{ad}, A_4) 
   \wedge \neg P_{\delta}(\textsf{ad}, A_5))))$,\\\\
    we have described that when  
     either of $A_4$ and $A_5$ is admissible in
    some reachable state,  
     there is always a branch where  
      the other becomes admissible and that becomes 
   not admissible in some future state, that is to say, 
     there can be an infinite number of oscillation among states 
that admit different sets of arguments. 
\end{example}   
Straightforwardly: 
\begin{proposition}[Non-monotonicity of admissibility] {\ }\\
    Suppose APA $(A, R, \Rp, A_0, \hookrightarrow)$, and suppose that 
   a 
   set of arguments $A_x \subseteq A$ 
    is admissible in a reachable state 
   $F(A_1)$. It is not necessary that $A_x$ be admissible 
     in a state $F(A_2)$ which satisfies  
     $F(A_1) \hookrightarrowR{A_y} F(A_2)$
     for some $A_y \subseteq A$. 
\end{proposition}
\section{Discussion and Related Work}   
For dynamics of argumentation, 
adaptation of the AGM-like belief revision 
\cite{Alchourron85,Katsuno92} to argumentation systems 
\cite{Baumann15,Coste-Marquis14,Coste-Marquis14b,Cayrol10,Cayrol11,Rotstein08} is popularly studied.  
In these studies, the focus is on 
restricting the class of resulting argumentation frameworks (post-states)
by means of postulates for a given 
argumentation-framework (pre-state) and some action (add/remove 
an argument/attack/argumentation framework). 
In APA, generation by inducement 
and modification by conversion 
are primarily defined. Removal of an argument, however, 
is easily emulated through conversion by 
setting $a_1 = a_3$ in 
$(a_1, a_2, a_3) \in \Rp$. In the literature of belief revision theory, 
some consider selective revision \cite{HanssonFCF01}, 
where a change to a belief set takes place {\it if 
the input that is attempting a change is accepted}. While 
such screening should be best assumed 
to have taken place beforehand within belief revision, 
a similar idea is critical in argumentation theory 
where defence of an argument is foundationally tied to a reference
 set of arguments.
Since any set of arguments may be chosen to be a reference set, 
and since which arguments in the set are visible 
non-monotonically changes, it is not feasible 
to assume some persuasion acts successful and others not 
in all states. \\
\indent Coordination of dynamics and statics
is somehow under-investigated in the literature 
of argumentation theory. A kind 
for coalition profitability and formability semantics 
with what are termed conflict-eliminable sets of arguments  
\cite{Arisaka17a} focuses on the interaction 
between sets of arguments before and after coalition formation.  
Doutre {\it et al.} show the use of propositional 
dynamic logic in program analysis/verification for 
encoding Dung theory and addition/removal 
of attacks and arguments \cite{Doutre14,Doutre17}. The 
logic comes 
with sequential operations, non-deterministic operations, tests. 
In comparison to their logic, our theory  
is an extension to Dung theory, which already 
provides a sound theoretical judgement for defence against 
attacks, which we extended also to persuasion acts. As far 
as we could fathom, such interaction between 
attack, persuasion, and defence has not been 
primarily studied 
in the literature. For another, 
a Dung-based theory has a certain appeal as a higher-level 
specification language.  Consider the argumentation 
in Figure \fbox{D}. 
APA requires 4 arguments, 
its subset as the set of initially visible arguments, 
2 inducements and 2 conversions for specification of 
the dynamic argumentation. By contrast,  
specification of a dynamic argumentation  
in the dynamic logic can be exponentially long 
as the number of non-deterministic branches increases; 
for the same dynamic argumentation in Figure \fbox{D}, 
it requires descriptions 
of all possible reachable states and transitions among them 
for the specification. 
We might 
take an analogy in chess here. While the number of 
branches in a chess game 
is astronomical, the game itself is specifiable in a small set 
of rules. For yet another, the dynamic logic facilitates 
dynamic changes to attacks in addition to arguments, 
which we did not study in this paper. The reason is mostly 
due to such consideration bound to lead to recursive persuasions and 
attacks (for 
recursive attacks/supports, see 
\cite{Barringer05,Gabbay09b,Arisaka16d,Baroni11,Cayrol18}) in our theory, 
which we believe will be better 
detailed in a separate paper for more formal interest. \\
\indent Argumentation theories that accommodate 
aspects of persuasion have been noted across several papers.  
In \cite{Bench-Capon03}, argumentation frameworks 
were augmented with values that controlled 
defeat judgement. Compared to their work, persuasion 
acts in APA are stand-alone relations which may be `executed' 
non-deterministically and concurrently, may \linebreak irreversibly modify 
visible arguments, and may produce loops. 
In most of argumentation papers on this topic, persuasion or negotiation 
is treated in a dialogue game \cite{Bench-Capon07b,Amgoud00,Black15,Fan11,Hadjinikolis13,Hunter15,McBurney03,Prakken05b,Prakken06,Rienstra13} where proponent(s) and opponent(s) take 
turns to modify an argumentation framework. 
APA does not assume the turn-based nature. 
In real-time rhetoric argumentation, as also frequently seen in social forums, 
more than one dialogue or more than one line of 
persuasive act may be running simultaneously. In 
this work, we were more interested
in modelling those situations. 
The various admissibility judgement enabled by (effectively) CTL (and 
other branching-time logic) should 
provide means of describing 
many types of argumentation queries. \\
\indent Studies on temporal arguments include 
\cite{Mann08,Barringer10,Augusto99,Budan17}. Most 
of these actually consider 
arguments that may be time-dependent. APA frameworks
keep arguments abstract, and observe 
temporal progress through actual execution of persuasive acts. 
We use 
temporal logic for describing admissibilities rather than   
arguments (recall that $P_\delta(\omega, A_x)$ is a formula 
on admissibility, not an argument).   
In timed argumentation frameworks \cite{Budan17}, 
arguments are available for set periods of time. Combined 
with APA, it should become possible to explain how and why 
arguments are available for the durations of time 
in the frameworks, the explanatory power incidentally having 
been  
the strength of argumentation theory. 
\section{Conclusion}       
We have shown a direction for 
abstract argumentation 
with dynamic operators extending Dung's theory. We 
set forth important properties and notions, and showed 
embedding of state-wise admissibility into CTL for 
various admissibilities across transitions. 
Many technical developments are expected to
follow. Our contribution 
is promising for bringing together knowledge of
abstract argumentation in AI and techniques 
and issues 
of concurrency in program analysis in a very near future. 
Cross-studies in the two domains are highly expected. Study in  
concurrent aspects of argumentation 
is important for evaluation of opinion transitions, 
which influences development of more effective 
sales approaches and better marketing in business, 
and consensus control tactics in politics. Harnessing  
our study with probabilistic methods is likely to form exciting
research. For future work, 
we plan to: take into account nuances of persuasive 
acts such as pseudo-logic, scapegoating, threat, and half-truths 
\cite{Chomsky10}; and extend APA with multi-reference sets.  
\section*{Acknowledgements} 
We thank anonymous reviewers for helpful comments.  
There was one suggestion concerning terms: to say to 
``convince'' instead of ``actively persuade'' or ``convert''. 
We seriously contemplated the suggested modification, 
and only 
in the end chose to leave the text as it was.  
%
\bibliography{references} 

\begin{thebibliography}{10}

\bibitem{Alchourron85}
Carlos~E. Alchourr{\'o}n and David Makinson.
\newblock {On the logic of theory change: Safe contraction}.
\newblock {\em {Studia Logica}}, 44:405--422, 1985.

\bibitem{Amgoud00}
Leila Amgoud, Simon Parsons, and Nicolas Maudet.
\newblock {Arguments, dialogue and negotiation}.
\newblock In {\em {ECAI}}, pages 338--342, 2000.

\bibitem{Arisaka16d}
Ryuta Arisaka and Ken Satoh.
\newblock {Voluntary Manslaughter? A Case Study with Meta-Argumentation with
  Supports}.
\newblock In {\em {JSAI-isAI Workshops}}, pages 241--252, 2016.

\bibitem{Arisaka17a}
Ryuta Arisaka and Ken Satoh.
\newblock {Coalition Formability Semantics with Conflict-Eliminable Sets of
  Arguments}.
\newblock In {\em {AAMAS}}, pages 1469--1471, 2017.

\bibitem{Augusto99}
Juan~Carlos Augusto and Guillermo~R. Simari.
\newblock {A Temporal Argumentative System}.
\newblock {\em {AI Commun.}}, 12(4):237--257, 1999.

\bibitem{Baroni11}
Pietro Baroni, Federico Cerutti, Massimiliano Giacomin, and Giovanni Guida.
\newblock {AFRA: Argumentation framework with recursive attacks}.
\newblock In {\em {International Journal of Approximate Reasoning}}, volume~52,
  pages 19--37, 2011.

\bibitem{Barringer10}
Howard Barringer and Dov~M. Gabbay.
\newblock {Modal and Tempral Argumentation Networks}.
\newblock In {\em {Time for Verification}}, pages 1--25. Springer, 2010.

\bibitem{Barringer05}
Howard Barringer, Dov~M. Gabbay, and John Woods.
\newblock {Temporal Dynamics of Argumentation Networks}.
\newblock In {\em {Mechanizing Mathematical Reasoning}}, pages 59--98, 2005.

\bibitem{Baumann15}
Ringo Baumann and Gerhard Brewka.
\newblock {AGM Meets Abstract Argumentation: Expansion and Revision for Dung
  Frameworks}.
\newblock In {\em {IJCAI}}, pages 2734--2740, 2015.

\bibitem{Bench-Capon03}
Trevor J.~M. Bench-Capon.
\newblock {Persuasion in Practial Argument Using Value-based Argumentation
  Frameworks}.
\newblock {\em {Journal of Logic and Computation}}, 13(3):429--448, 2003.

\bibitem{Bench-Capon07b}
Trevor J.~M. Bench-Capon, Sylvie Doutre, and Paul~E. Dunne.
\newblock {Audiences in argumentation frameworks}.
\newblock {\em {Artificial Intelligence}}, 171(1):42--71, 2007.

\bibitem{Black15}
Elizabeth Black and Anthony Hunter.
\newblock {Reasons and Options for Updating an Opponent Model in Persuasion
  Dialogues}.
\newblock In {\em {TAFA}}, pages 21--39, 2015.

\bibitem{Budan17}
Maximiliano C.~D. Bud{\'a}n, Maria~Laura Cobo, Diego~C. Martinez, and
  Guillermo~R. Simari.
\newblock {Bipolarity in temporal argumentation frameworks}.
\newblock {\em {International Journal of Approximate Reasoning}}, 84:1--22,
  2017.

\bibitem{Cayrol18}
Caludette Cayrol, Jorge Fandinno, Luis~Fari{\~n}as del Cerro, and
  Marie-Christine Lagasquie-Schiex.
\newblock {Argumentation Frameworks with Recursive Attacks and Evidence-Based
  Supports}.
\newblock In {\em {FoIKS}}, pages 150--169, 2018.

\bibitem{Cayrol10}
Claudette Cayrol, Florence~Dupin de~Saint-Cyr, and Marie-Christine
  Lagasquie-Schiex.
\newblock {Change in Abstract Argumentation Frameworks: Adding an Argument}.
\newblock {\em {Journal of Artificial Intelligence Research}}, 38:49--84, 2010.

\bibitem{Cayrol11}
Claudette Cayrol and Marie-Christine Lagasquie-Schiex.
\newblock {Bipolarity in Argumentation Graphs: Towards a Better Understanding}.
\newblock In {\em {Scalable Uncertainty Management}}, pages 137--148, 2011.

\bibitem{Chomsky10}
Noam Chomsky.
\newblock {\em {Hopes and Prospects}}.
\newblock {Haymarket Books}, 2010.

\bibitem{Coste-Marquis14b}
Sylvie Coste-Marquis and S{\'e}bastien Konieczny.
\newblock {A Translation-Based Approach for Revision of Argumentation
  Frameworks}.
\newblock In {\em {JELIA}}, pages 397--411, 2014.

\bibitem{Coste-Marquis14}
Sylvie Coste-Marquis, S{\'e}bastien Konieczny, Jean-Guy Mailly, and Pierre
  Marquis.
\newblock {On the Revision of Argumentation Systems: Minimal Change of
  arguments Statuses}.
\newblock In {\em {KR}}, 2014.

\bibitem{Doutre14}
Sylvie Doutre, Andreas Herzig, and Laurent Perrussel.
\newblock {A Dynamic Logic Framework for Abstract Argumentation}.
\newblock In {\em {KR}}, 2014.

\bibitem{Doutre17}
Sylvie Doutre, Faustine Maffre, and Peter McBurney.
\newblock {A Dynamic Logic Framework for Abstract Argumentation: Adding and
  Removing Arguments}.
\newblock In {\em {IEA/AIE}}, pages 295--305, 2017.

\bibitem{Dung95}
Phan~M. Dung.
\newblock On the {Acceptability} of {Arguments} and {Its} {Fundamental} {Role}
  in {Nonmonotonic} {Reasoning}, {Logic Programming}, and n-{Person} {Games}.
\newblock {\em Artificial {Intelligence}}, 77(2):321--357, 1995.

\bibitem{Fan11}
Xiuyi Fan and Francesca Toni.
\newblock {Assumption-Based Argumentation Dialogues}.
\newblock In {\em {IJCAI}}, pages 198--203, 2011.

\bibitem{Gabbay09b}
Dov~M. Gabbay.
\newblock {Semantics for Higher Level Attacks in Extended Argumentation Frames
  Part 1: Overview}.
\newblock {\em {Studia Logica}}, 93(2-3):357--381, 2009.

\bibitem{Hadjinikolis13}
Christos Hadjinikolis, Yiannis Siantos, Sanjay Modgil, Elizabeth Black, and
  Peter McBurney.
\newblock {Opponent modelling in persuasion dialogues}.
\newblock In {\em {IJCAI}}, pages 164--170, 2013.

\bibitem{HanssonFCF01}
Sven~Ove Hansson, Eduardo~L. Ferm{\'e}, John Cantwell, and Marcelo~A. Falappa.
\newblock {Credibility Limited Revision}.
\newblock {\em {Journal of Symbolic Logic}}, 66(4):1581--1596, 2001.

\bibitem{Hunter15}
Anthony Hunter.
\newblock {Modelling the persuadee in asymmetric argumentation dialogues for
  persuasion}.
\newblock In {\em {IJCAI}}, pages 3055--3061, 2015.

\bibitem{Katsuno92}
H.~Katsuno and Alberto~O. Mendelzon.
\newblock {On the Difference between Updating a Knowledge Base and Revising
  it}.
\newblock In {\em {Belief Revision}}. Cambridge University Press, 1992.

\bibitem{Mann08}
Nicholas Mann and Anthony Hunter.
\newblock {Argumentation Using Temporal Knowledge}.
\newblock In {\em {COMMA}}, pages 204--215, 2008.

\bibitem{McBurney03}
Peter McBurney, Rogier van Eijk, Simon Parsons, and Leila Amgoud.
\newblock {A dialogue-game protocol for agent purchase negotiations}.
\newblock {\em {Journal of Autonomous Agents and Multi-Agent Systems}},
  7:235--273, 2003.

\bibitem{Prakken05b}
Henry Prakken.
\newblock {Coherence and Flexibility in Dialogue Games for Argumentation}.
\newblock {\em {J. Log. Comput.}}, 15(6):1009--1040, 2005.

\bibitem{Prakken06}
Henry Prakken.
\newblock {Formal Systems for Persuasion Dialogue}.
\newblock {\em {Knowledge Engineering Review}}, 21(2):163--188, 2006.

\bibitem{Rienstra13}
Tjitze Rienstra, Matthias Thimm, and Nir Oren.
\newblock {Opponent Models with Uncertainty for Stragegic Argumentation}.
\newblock In {\em {IJCAI}}, pages 332--338, 2013.

\bibitem{Rotstein08}
Nicol{\'a}s Rotstein, Mart{\'\i}n~O. Moguillansky, Alejandro~J. Garc{\'\i}a,
  and Guillermo~R. Simari.
\newblock {An abstract Argumentation Framework for Handling Dynamics}.
\newblock In {\em {Proceedings of the Argument, Dialogue and Decision Workshop
  in NMR 2008}}, pages 131--139, 2008.

\end{thebibliography}
\end{document}